\RequirePackage{amsmath}
\documentclass[runningheads,a4paper]{llncs}

\usepackage[utf8]{inputenc}
\usepackage[T2A]{fontenc}

\usepackage[russian,english]{babel}
\babeltags{ru=russian}

\usepackage{todonotes}
\usepackage{amsfonts,amssymb}
\usepackage{graphicx}
\usepackage[caption=false]{subfig}
\usepackage{cite}

\usepackage{array}
\newcolumntype{R}[1]{>{\raggedleft\let\newline\\\arraybackslash\hspace{0pt}}m{#1}}

\def\watset#{\textsc{Watset}}

\def\maxmax#{\textsc{MaxMax}}

\definecolor{ggplotoriginal}{HTML}{238b45}
\definecolor{ggplottransitive}{HTML}{2171b5}
\definecolor{ggplotmerging}{HTML}{cb181d}

\title{Fighting with the Sparsity of Synonymy Dictionaries for Automatic Synset Induction}

\titlerunning{Fighting with the Sparsity of Synonymy Dictionaries}

\author{
  Dmitry Ustalov\inst{1,2} \and
  Mikhail Chernoskutov\inst{1,2} \and
  Chris Biemann\inst{3} \and \\
  Alexander Panchenko\inst{3}
}

\authorrunning{Dmitry Ustalov et al.}

\institute{
  Ural Federal University, Yekaterinburg, Russia
  \and
  Krasovskii Institute of Mathematics and Mechanics, Yekaterinburg, Russia\\
  \email{\{dmitry.ustalov,mikhail.chernoskutov\}@urfu.ru}
  \and
  Universit\"{a}t Hamburg, Hamburg, Germany\\
  \email{\{biemann,panchenko\}@informatik.uni-hamburg.de}
}

\begin{document}

\maketitle

\begin{abstract} 
Graph-based synset induction methods, such as \maxmax{} and \watset{}, induce synsets by performing a global clustering of a synonymy graph. However, such methods are sensitive to the structure of the input synonymy graph: sparseness of the input dictionary can substantially reduce the quality of the extracted synsets. In this paper, we propose two different approaches designed to alleviate the incompleteness of the input dictionaries. The first one performs a pre-processing of the graph by adding missing edges, while the second one performs a post-processing by merging similar synset clusters. We evaluate these approaches on two datasets for the Russian language and discuss their impact on the performance of synset induction methods. Finally, we perform an extensive error analysis of each approach and discuss prominent alternative methods for coping with the problem of sparsity of the synonymy dictionaries.

\end{abstract}

\section{Introduction}

A synonymy dictionary, representing synonymy relations between the individual words, can be modeled as an undirected graph where nodes are words and edges are synonymy relations.\footnote{In the context of this work, we assume that synonymy is a relation of lexical semantic equivalence which is context-independent, as opposed to ``contextual synonyms''~\cite{zeng2007semantic}.} Such a graph, called a synonymy graph or a synonymy network, tends to have a clustered structure~\cite{Gfeller:05}. This property is exploited by various graph-based word sense induction (WSI) methods, such as~\cite{Panchenko:16}. The goal of such WSI methods is to build a word sense inventory from various networks, such as synonymy graphs, co-occurrence graphs, graphs of distributionally related words, etc. (see a survey by Navigli~\cite{Navigli:12}). 

The clusters are densely connected subgraphs of synonymy graph that correspond to the groups of semantically equivalent words or synsets (sets of synonyms). Synsets are building blocks for WordNet~\cite{Fellbaum:98} and similar lexical databases used in various applications, such as information retrieval~\cite{Loukachevitch:11}. Graph-based WSI combined with graph clustering makes it possible to induce synsets in an unsupervised way~\cite{Hope:13,Ustalov:17:acl}. However, these methods are highly sensitive to the structure of the input synonymy graph~\cite{Ustalov:17:acl}, which motivates the development of synonymy graph expansion methods. 

In this paper, we are focused on the data sparseness reduction problem in the synonymy graphs. This problem is inherent to the majority of manually constructed lexical-semantic graphs due to the Zipf's law of word frequencies~\cite{zipf1935psychobiology}: the long tail of rare words is inherently underrepresented. In this work, given a synonymy graph and a graph clustering algorithm, we compare the performance of two methods designed to improve synset induction. The goal of each method is  to improve the final synset cluster structures. Both methods are based on the assumption that synonymy is a symmetric relation. We run our experiments on the Russian language using the \watset{} state-of-the-art unsupervised synset induction method~\cite{Ustalov:17:acl}. 

The contribution of this paper is a study of two principally different methods for dealing with the sparsity of the input synonymy graphs. The former, \textit{relation transitivity} method, is based on expansion of the synonymy graph. The latter, \textit{synset merging} method, is based on the mutual similarity of synsets. 

\section{Related Work}

Hope and Keller \cite{Hope:13} introduced the \maxmax{} clustering algorithm particularly designed for the word sense induction task. In a nutshell, pairs of nodes are grouped if they have a maximal mutual affinity. The algorithm starts by converting the undirected input graph into a directed graph by keeping the maximal affinity nodes of each node. Next, all nodes are marked as root nodes. Finally, for each root node, the following procedure is repeated: all transitive children of this root form a cluster and the root are marked as non-root nodes; a root node together with all its transitive children form a fuzzy cluster.

Van~Dongen\cite{vanDongen:00} presented the Markov Clustering (MCL) algorithm for graphs based on simulation of stochastic flow in graphs. MCL simulates random walks within a graph by alternation of two operators called expansion and inflation, which recompute the class labels. This approch has been successfully used for the word sense induction task~\cite{Dorow:03}.

Biemann \cite{Biemann:06} introduced Chinese Whispers, a clustering algorithm for weighted graphs that can be considered as a special case of MCL with a simplified class update step. At each iteration, the labels of all the nodes are updated according to the majority labels among the neighboring nodes. The author showed usefulness of the algorithm for induction of word senses based on corpus-induced graphs. 

The ECO approach~\cite{GoncaloOliveira:14} was applied to induce a WordNet of the Portuguese language.\footnote{\url{http://ontopt.dei.uc.pt}} 
In its core, ECO is based on a clustering algorithm that was used to induce synsets from synonymy dictionaries. The algorithm starts by adding random noise to edge weights. Then, the approach applies Markov Clustering of this graph several times to estimate the probability of each word pair being in the same synset. Finally, candidate pairs over a certain threshold are added to output synsets.

In our experiments, we rely on the \watset{} synset induction method~\cite{Ustalov:17:acl} based on a graph meta-clustering algorithm that combines local and global hard clustering to obtain a fuzzy graph clustering. The authors shown that this approach outperforms all methods mentioned above on the synset induction task and therefore we use it as the strongest baseline to date.

Meyer and Gurevich \cite{Meyer:12} presented an approach for construction of an ontologized version of Wiktionary, by formation of ontological concepts and relationships between them from the ambiguous input dictionary, yet their approach does not involve graph clustering.

\section{Two Approaches to Cope with Dictionary Sparseness}\label{sec:approach}

We propose two approaches for dealing with the incompleteness of the input synonymy dictionaries of a  graph-based synset induction method, such as \watset{} or \maxmax{}. First, we describe a graph-based approach that preprocesses the input graph by adding new edges. This step is applied \textit{before} the synset induction clustering. Second, we describe an approach that post-processes the synsets by merging highly semantically related synsets. This step is applied \textit{after} the synset induction clustering step, refining its results. 

\subsection{Expansion of Synonymy Graph via Relation Transitivity}

Assuming that synonymy is an equivalence relation due to its reflexiveness, symmetry, and transitivity, we can insert additional edges into the synonymy graph between nodes that are transitively, synonymous, i.e. are connected by a short path of synonymy links. 
We assume that if an edge for a pair of synonyms is missing, the graph still contains several relatively short paths connecting the nodes corresponding to these words.

Firstly, for each vertex, we extract its neighbors and the neighbors of these neighbors. Secondly, we compute the set of candidate edges by connecting the disconnected vertices. Then, we compute the number of simple paths between the vertices in candidate edges. Finally, we add an edge into the graph if there are at least $k$ such paths which lengths are in the range $[i, j]$.

Particularly, the algorithm works as follows:
\begin{enumerate}
  \item extract a first-order ego network $N_1$ and a second-order ego network $N_2$ for each node;
  \item generate the set of candidate edges that connect the disconnected nodes in $N_1$, i.e., the total number of the candidates is $C_{|N_1|}^2 - |E_{N_1}|$, where $C_{|N_1|}^2$ is the number of all 2-combinations over the $|N_1|$-element set and $E_{N_1}$ is the set of edges in $N_1$;
  \item keep only those edge candidates that satisfy two conditions: 1) there are at least $k$ paths $p$ in $N_2$, so no path contains the initial ego node, and 2) the length of each path belongs to the interval $[i, j]$.
\end{enumerate}

The approach has two parameters: the minimal number of paths to consider $k$ and the path length interval $[i, j]$. It should be noted that this approach processes the input synonymy graph without taking the polysemous words into account. Such words are then handled by the \watset{} algorithm that induces word senses based on the expanded synonymy graph. 

\subsection{Synset Merging based on Synset Vector Representations}

We assume that closely related synsets carry equivalent meanings and use the following procedure to merge near-duplicate synsets:
\begin{enumerate}
  \item learn synset embeddings for each synset using the SenseGram method by simply averaging word vectors that correspond to the words in the synset~\cite{Pelevina:16};
  \item identify the closely related synsets using the $m\text{-}k\textit{NN}$ algorithm~\cite{Panchenko:12:cdud} that considers two objects as closely related if they are mutual neighbors of each other;
  \item merge the closely related synsets in a specific order: the smallest synsets are merged first, the largest are merged later; every synset can be merged only once in order to avoid giant merged clusters.
\end{enumerate}

This approach has two parameters: the number of nearest neighbors to consider $k$ (fixed to $10$ in our experiments)\footnote{In general, the $m\text{-}k\textit{NN}$ method can be parametrized by two different parameters: $k_{ij}$ -- the number of nearest neighbors from the word $i$ to the word $j$ and $k_{ji}$ -- the number of nearest neighbors from the word $j$ to the word $i$. In our case, for simplicity, we set $k_{ij} = k_{ji} = k$. } and the maximal number of merged synsets $t$, e.g., if $t=1$ then only the first mutual nearest neighbor is merged. It should be noted that this approach operates on synsets that which are already have been discovered by \watset{}. Therefore, the merged synsets are composed of disambiguated word senses.

\section{Evaluation}\label{sec:experiments}

We evaluate the performance of the proposed approaches using the \watset{} graph clustering method that shows state-of-the-art results on synset induction~\cite{Ustalov:17:acl}. \watset{} is a meta-clustering algorithm that disambiguates a (word) graph by first performing ego-network clustering to split nodes (words) into (word) senses. Then a global clustering is used to form (syn)sets of senses. For both clustering steps, any graph clustering algorithm can be employed; in~\cite{Ustalov:17:acl}, it was shown that combinations of Chinese Whispers~\cite{Biemann:06} (CW) and Markov Clustering~\cite{vanDongen:00} (MCL) provide the best results. We also evaluated the same approaches with the \maxmax{}~\cite{Hope:13} method, but the results were virtually the same, so we omitted them for brevity.

\subsection{Datasets}

We evaluate the proposed augmentation approaches on two gold standard datasets for Russian: RuWordNet~\cite{Loukachevitch:16} and YARN~\cite{Braslavski:16}. Both are analogues of the original English WordNet~\cite{Fellbaum:98}. 

We used the same input graph as in~\cite{Ustalov:17:acl}; the graph is based on three synonymy dictionaries, the Russian Wiktionary, the Abramov's dictionary and the UNLDC dictionary. The graph is weighted using the similarities from Russian Distributional Thesaurus (RDT)~\cite{Panchenko:17}.\footnote{\url{http://russe.nlpub.ru/downloads}}. To construct synset embeddings, we used word vectors from the RDT. 

The lexicon of the input dictionary is different from the lexicon of RuWordNet~\cite{Loukachevitch:16}, which includes a lot of domain-specific synsets. At the same time, the input dataset is the same as the data sources used for boostrapping YARN~\cite{Braslavski:16}.

The summary of the datasets is shown in \tablename~\ref{tab:datasets}: the ``\#~words'' column specifies the number of lexical units in the dataset (nodes of the input graph), the ``\#~synonyms'' column indicates the number of synonymy pairs appearing in the dataset (edges of the input graph). The problem of dictionary sparsity is the fact that some edges (synonyms) are missing in the input resource. Finally, the ``\#~synsets'' column specifies the number of resulting synsets (if applicable).

\begin{table}[ht]
\centering
\caption{Summary of the datasets used in the experiments.}
\begin{tabular}{p{60mm}*{3}{|R{20mm}}}
\textbf{Resource} & \textbf{\#~words} & \textbf{\#~synsets} & \textbf{\#~synonyms} \\\hline
Input Synonymy Dictionary: Wiktionary &      $83\,092$ &           n/a &     $211\,986$ \\\hline
Induced Synsets: \watset{} MCL-MCL   &      $83\,092$ &     $36\,217$ &     $406\,430$ \\
Induced Synsets: \watset{} CW-MCL    &      $83\,092$ &     $55\,369$ &     $355\,158$ \\\hline
Gold Synsets: RuWordNet        &     $110\,242$ &     $49\,492$ &     $278\,381$ \\
Gold Synsets: YARN             &       $9\,141$ &      $2\,210$ &      $48\,291$ \\
\end{tabular}
\label{tab:datasets}
\end{table}

\subsection{Quality Measures}

We report results according to standard word sense induction evaluation measures: paired precision, recall and F-score~\cite{Manandhar:10}, i.e., each cluster of $n$ words yields $\frac{n(n-1)}{2}$ synonymy pairs. The exact same evaluation protocol was used in the original \watset{} publication. We perform evaluation on the intersection of gold standard lexicon and the lexicon of the induced resource. 

\subsection{Results}

\begin{figure}
\centering
\subfloat[RuWordNet]{\includegraphics[width=.4725\textwidth]{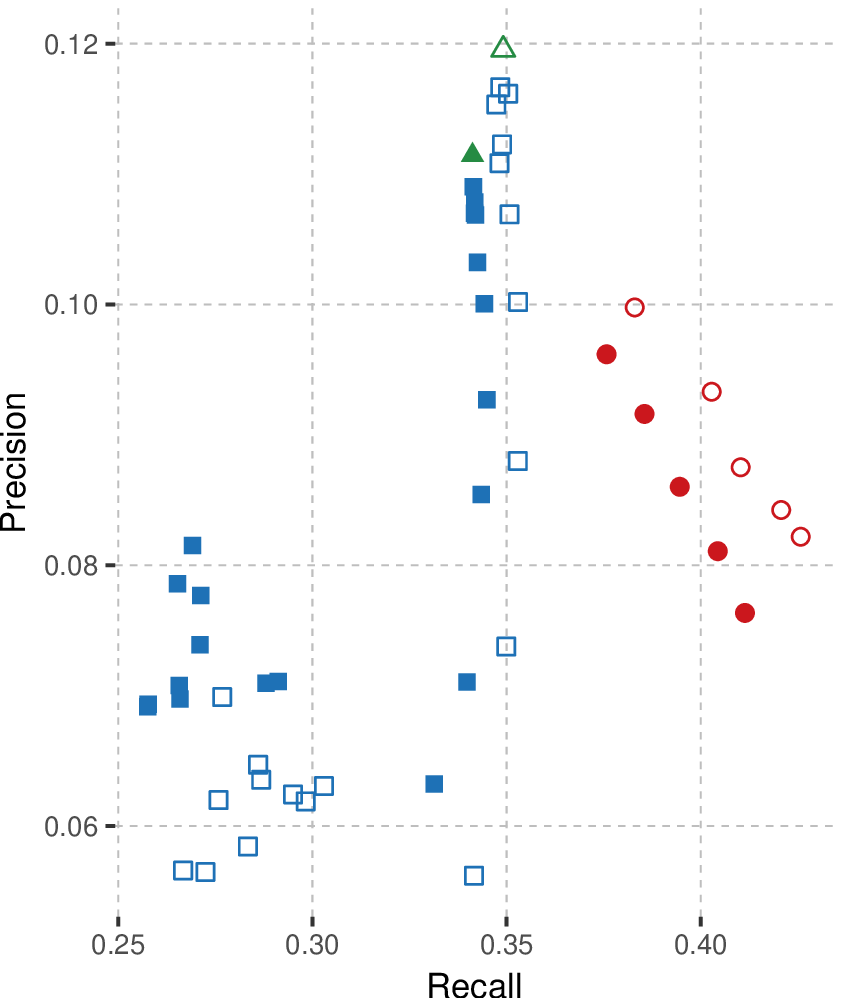}}
\qquad
\subfloat[Yet Another RussNet]{\includegraphics[width=.4725\textwidth]{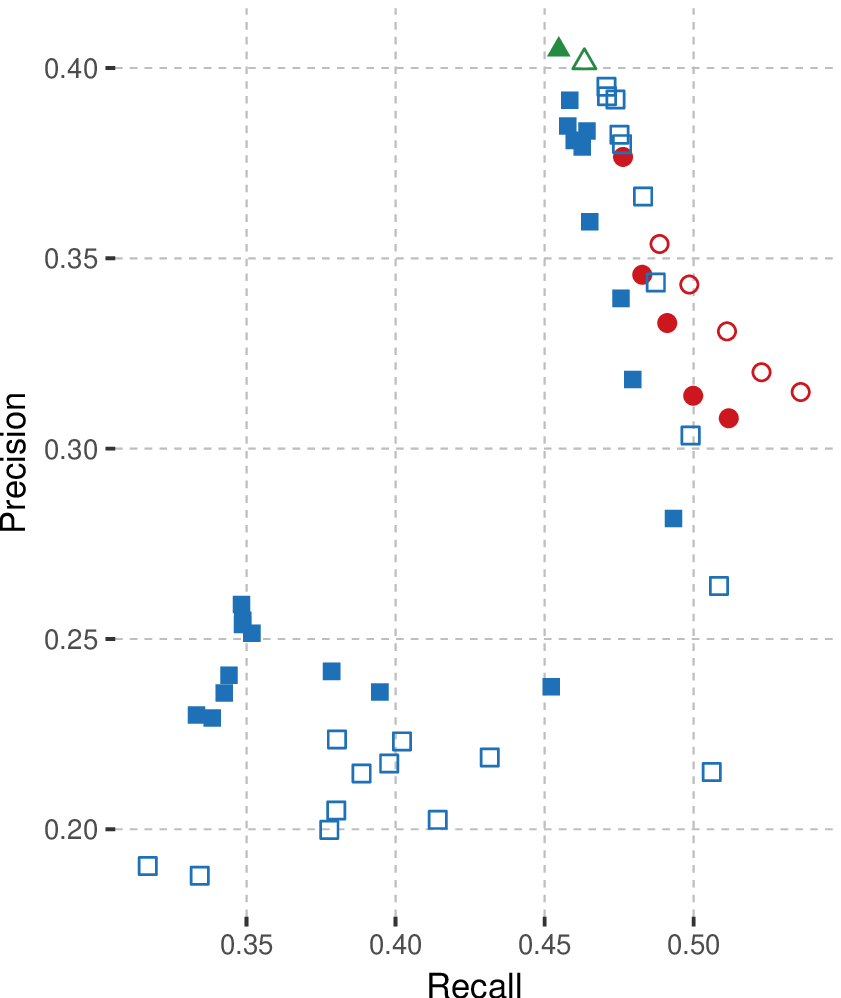}}
\caption{Precision-recall plots built on two gold standard datasets for Russian. The shapes and colors: $\color{ggplotoriginal}\blacktriangle$\,original graph, $\color{ggplottransitive}\blacksquare$\,transitivity expansion with the allowed path lengths $2 \leq i \leq j \leq 3$ and the number of simple paths $k \leq 10$, $\color{ggplotmerging}\bullet$\,synset merging with the maximal number of merged mutual neighbours $t \in \{1,2,3,5,10\}$ and the number of nearest neighbours set to $k=10$. Chinese Whispers-based \watset{} configurations are hollow, while MCL-based are solid. }
\label{fig:plot}
\end{figure}

The evaluation results are shown in \figurename~\ref{fig:plot}. As one may observe, in the case of the RuWordNet dataset, the method based on the transitivity expansion rendered almost no improvements in terms of recall while dramatically dropping the precision. The second method, based on synset embeddings shows much better results on this dataset: It substantially improves recall, yet at the cost of a drop in precision. 

In case of the YARN dataset, the results are similar with the graph-based method significantly lagging behind the vector-based method. However, in this case, the difference in the observed performance is smaller with some configurations of the graph-based methods approaching the performance of the vector-based method. Similarly to the first dataset, both methods trade off gains in recall for the drops in precision. Note, however, that the vector-based method can perform a shift of the ``sweet spot'' of the clustering approach. While the F-measure remains at the same level, it is possible to obtain higher levels of recall, which can be useful for some applications. In the following section, we perform error analysis for each method. 

\section{Discussion}\label{sec:discussion}

Perfect synonyms are very rare, which is confirmed by the precision-recall plot in \figurename~\ref{fig:plot}. Both methods insert relations of other types, such as association, co-hyponymy, hypernymy, etc. Recall increases with the level of inclusiveness of the configuration; this also causes significant drops in precision. The expansion methods presented in this paper could  therefore be more useful for generation of other types of symmetric semantic relations, such as co-hyponymy.

\subsection{Error Analysis: Synonymy Transitivity}

We tried the following configurations of the approach: $2 \leq i \leq j \leq 3$, $k \leq 10$. However, only the variations with a small allowed length $i = j = 2$ and a high number of found simple paths $k \geq 5$ yielded viable results.

We explain the quick drops in precision by the fact that no word is a perfect synonym of another~\cite{Herrmann:78}. This results in the potential loss of the synonymy relation on each additional transitive node. While having a lot of pertinent edge insertions like ``\textru{оказываться}--\textru{появляться} (show up -- appear)'' or ``\textru{подтрунивать}--\textru{стебаться} (prank -- make jokes)'', this method introduces such false positives like ``\textru{кий}--\textru{хлыст} (cue -- whip)'', ``\textru{шеф}--\textru{царь} (boss -- tsar)'', ``\textru{солидный}--\textru{корректный} (solid -- correct)'', etc.

One of the reasons of this outcome is that adding new edges increases the size of the communities. They capture neighboring vertices and edges belonging to other communities in the initial graph. Hence, on the one hand, we obtain communities with excess elements, while on the other hand, we observe depleted communities.

\subsection{Error Analysis: Synset Merging}

The different configurations of the vector-based method in this plot correspond to the following values of the $t$ parameter (the maximum number of merged synsets): 1, 2, 3, 5, 10. Merging more than one synset at a time provides a substantial gain in the recall, yet again at the cost of the precision drop.

\tablename~\ref{tab:correct} presents an example of correct merging of synsets. The results of the clustering generate multiple small synsets that refer to the same meaning. Such synsets tend to be mutual nearest neighbors. Top ten most similar synsets to the synset ``cynicism'' are depicted. In this table, we also indicate whether each neighbor is the mutual nearest neighbor or not. In this example, the method of mutual nearest neighbors perfectly achieves its goal of merging synonymous synsets.

\tablename~\ref{tab:wrong} presents an example of a wrong merging of synsets on the example of the synset ``zinc, Zn''. This sample illustrates the reasons behind the drops in precision. While different chemical elements, such as zinc and cobalt are strongly semantically related, they are co-hyponyms of the common hypernym ``chemical element'', and not synonyms, i.e. terms with equivalent meanings. This result is in line with the prior results showing that the majority of the nearest neighbors delivered by the distributional semantic models, such as the skip-gram model~\cite{Mikolov:13} used in our experiments, tend to be co-hyponyms as shown in prior studies~\cite{Wandmacher:05,Heylen:08,Peirsman:08,panchenko2011comparison}. The results presented in both \tablename~\ref{tab:correct} and \ref{tab:wrong} have been manually annotated by a single expert.

\begin{table}
\centering
\caption{An example of correct synset merging, where predicted labels are equal to gold labels for the top $k=10$ nearest neighbours of the synset ``цинизм, циничность (cynicism, cynicism)''. In this table, the ``Predicted'' column contains mutually related synsets, while the ``Gold'' column lists expert judgments.}
\begin{tabular}{c|c|p{73mm}|c|c}
$k$ & \textbf{Similarity} & \textbf{Related Synset} & \textbf{Predicted} & \textbf{Gold} \\ \hline
1 & 0.866 & \textru{беспринципность, цинизм} (unprincipledness, cynicism) & true & true \\ \hline
2 & 0.856 & \textru{беспринципность, циничность} (unprincipledness, cynicism) & true & true  \\ \hline
3 & 0.853 & \textru{кинизм, беспардонность, цинизм} (cynicism, shamelessness, cynicism) & true & true  \\ \hline
4 & 0.734 & \textru{нахрапистость, нахальство, нахальность, циничность, бесцеремонность, нецеремонность} (сheekiness, impudence, cheekiness, cynicism, brusqueness, unceremoniousness) & false & false \\ \hline
5 & 0.677 & \textru{грубость, примитивизм} (rudeness, primitivism) & false & false \\ \hline
6 & 0.677 & \textru{хамство, лапидарность, хамёж, топорность, грубость, прямолинейность} (rudeness, conciseness, rudeness, clumsiness, rudeness, straightness) & false & false \\ \hline
7 & 0.674 & \textru{безнравственность, беспринципность, злонравие, аморальность} (wickedness, lack of principles, depravity, immorality) & false & false \\ \hline
8 & 0.671 & \textru{бесстыдство, непристойность, бессовестность, нахрап} (immorality, lack of principle, malice, immorality) & false & false \\ \hline
9 & 0.663 & \textru{скепсис, скептичность} (skepticism, skepticism) & true & false  \\ \hline
10 & 0.661 & \textru{фанатизм, ханжество} (bigotry, bigotry) &    false & false  \\
\end{tabular}
\label{tab:correct}
\end{table}

\begin{table}
\centering
\caption{An example of wrong synset merging, where predicted labels are not equal to gold labels for the top $k=10$ nearest neighbours of the synset ``цинка, Zn (zinc, Zn)''. In this table, the ``Predicted'' column contains mutually related synsets, while the ``Gold'' column lists expert judgments }
\begin{tabular}{c|c|p{73mm}|c|c}
$k$ & \textbf{Similarity} & \textbf{Related Synset} & \textbf{Predicted} & \textbf{Gold} \\ \hline
1 & 0.676 & \textru{станнат кобальта, кобальт} (cobalt stannate, cobalt) & true & false \\ \hline
2 & 0.673 & \textru{Mg, магний} (Mg, magnesium) & true  & false \\ \hline
3 & 0.670 & \textru{глиний, крылатый металл, алюминий, Al} (сlay, winged metal, aluminum, Al) & false  & false \\ \hline
4 & 0.663 & \textru{фосфор, P} (phosphorus, P) & true  & false\\ \hline
5 & 0.646 & \textru{оксид, окись} (oxide, oxide) & false  & false \\ \hline
6 & 0.631 & \textru{гидрооксид, гидроокись, гидроксид} (hydroxide, hydroxide, hydroxide) & false  & false \\ \hline
7 & 0.630 & \textru{ванадий, V} (vanadium, V) & true  & false \\ \hline
8 & 0.628 & \textru{рибофлавин, лактофлавин, витамин В} (riboflavin, lactoflavin, vitamin B) & false  & false \\ \hline
9 & 0.624 & \textru{иодат, йодид, йодат} (iodate, iodide, iodate) & false  & false \\ \hline
10 & 0.618 & \textru{кремний, Si} (silicon, Si) & false  & false \\
\end{tabular}
\label{tab:wrong}
\end{table}

\subsection{Other Ways to Deal with Sparseness of the Input Dictionary}

In this section, we discuss avenues for future work: the prominent approaches that might be useful in addressing the sparseness of the synonym dictionaries.

\paragraph{Lexical-syntactic patterns for extraction of synonyms.} Hearst patterns are widely used to mine hypernymy relations from text~\cite{Seitner:16}. Such patterns can be also learned automatically~\cite{Snow:04,Shwartz:16}. In \cite{Panchenko:12:konvens}, seven patterns for extraction of synonyms were proposed, which function in the same was as the Hearst patterns for hypernymy extraction. Such synonymy extraction patterns can be also learned automatically in the same fashion as patterns for hypernymy extraction from text~\cite{Snow:04}. Finally, hypernyms, antonyms, and other relations extracted from text can be used to filter our non-synonymous candidates.

\paragraph{Global clustering of synsets.} It is possible to find groups of semantically related words (clique-like structures) and expand synonyms only within such communities with a graph clustering algorithm, such as Chinese Whispers~\cite{Biemann:06}. The current graph-based transitivity expansion method does not consider the structure of the communities of the synonymy graph.

\paragraph{Synonymy detection as an anaphora resolution problem.} Another observation is that synonyms are often not used in the same sentence, but instead used to ensure linguistic variance of the text. In this respect, synonymy extraction task is similar to the anaphora resolution task~\cite{Lappin:94}. This line of work is related to prior work of~\cite{Feuerbach:15} in detecting  ``bridging mentions''.

\paragraph{Crowdsourcing.} Finally, the last option is simply to improve the quality of the input dictionaries by the means of crowdsourcing~\cite{Gurevych:13}. Namely, involving more people to edit Wiktionary that we use as the input data will increase the coverage of the extracted synsets, but large-scale crowdsourcing requires a set of elaborated quality control measures.  

\section{Conclusion}\label{sec:conclusion}

In this paper, we explored two alternative strategies for coping with the problem of inherent sparsity and incompleteness of the synonymy dictionaries. These sparsity issues hamper performance of the methods for automatic induction of synsets, such as \maxmax{}~\cite{Hope:13} and \watset{}~\cite{Ustalov:17:acl}. One of the proposed methods performs pre-processing of the graph of synonyms, while the second one performs post-processing of the induced synsets.

Our experiments on two large scale datasets show that (1) both methods are able to substantially improve recall, but at the cost of substantial drops of precision; (2) the post-processing approach yields better results overall. We conclude our study with an overview of prominent alternative approaches for expansion of incomplete synonymy dictionaries.

We believe the results of our study will be useful for both enriching the available lexical semantic resources like OntoWiktionary~\cite{Meyer:12} as well as for increasing the lexical coverage of the input data for the graph-based word sense induction methods.

\subsubsection*{Acknowledgements.} We acknowledge the support of the Deutsche Forschungsgemeinschaft (DFG) under the ``JOIN-T'' project, the DAAD, the RFBR under the projects no.~16-37-00203~\textru{мол\_а} and no.~16-37-00354~\textru{мол\_а}, and the RFH under the project no.~16-04-12019. The calculations were carried out using the supercomputer ``Uran'' at the Krasovskii Institute of Mathematics and Mechanics. Finally, we also thank four anonymous reviewers for their helpful comments.

\bibliographystyle{splncs03}
\bibliography{aug.aist2017}

\begin{thebibliography}{10}
\providecommand{\url}[1]{\texttt{#1}}
\providecommand{\urlprefix}{URL }

\bibitem{Biemann:06}
Biemann, C.: {Chinese Whispers: An Efficient Graph Clustering Algorithm and Its
  Application to Natural Language Processing Problems}. In: Proceedings of the
  First Workshop on Graph Based Methods for Natural Language Processing. pp.
  73--80. TextGraphs-1, Association for Computational Linguistics, New York,
  NY, USA (2006)

\bibitem{Braslavski:16}
Braslavski, P., Ustalov, D., Mukhin, M., Kiselev, Y.: {YARN:
  Spinning-in-Progress}. In: Proceedings of the 8th Global WordNet Conference.
  pp. 58--65. GWC 2016, Global WordNet Association, Bucharest, Romania (2016)

\bibitem{vanDongen:00}
van Dongen, S.: {Graph Clustering by Flow Simulation}. Ph.D. thesis, University
  of Utrecht (2000)

\bibitem{Dorow:03}
Dorow, B., Widdows, D.: {Discovering Corpus-Specific Word Senses}. In:
  Proceedings of the Tenth Conference on European Chapter of the Association
  for Computational Linguistics - Volume 2. pp. 79--82. EACL '03, Association
  for Computational Linguistics, Budapest, Hungary (2003)

\bibitem{Fellbaum:98}
Fellbaum, C.: {WordNet: An Electronic Database}. MIT Press (1998)

\bibitem{Feuerbach:15}
Feuerbach, T., Riedl, M., Biemann, C.: {Distributional Semantics for Resolving
  Bridging Mentions}. In: Proceedings of the International Conference Recent
  Advances in Natural Language Processing. pp. 192--199. INCOMA Ltd. Shoumen,
  BULGARIA, Hissar, Bulgaria (2015)

\bibitem{Gfeller:05}
Gfeller, D., Chappelier, J.C., De~Los~Rios, P.: {Synonym Dictionary Improvement
  through Markov Clustering and Clustering Stability}. In: Proceedings of the
  International Symposium on Applied Stochastic Models and Data Analysis. pp.
  106--113 (2005)

\bibitem{GoncaloOliveira:14}
Gon{\c{c}}alo~Oliveira, H., Gomes, P.: {ECO and Onto.PT: a flexible approach
  for creating a Portuguese wordnet automatically}. Language Resources and
  Evaluation  48(2),  373--393 (2014)

\bibitem{Gurevych:13}
Gurevych, I., Kim, J. (eds.): {The People's Web Meets NLP}. Springer Berlin
  Heidelberg (2013)

\bibitem{Herrmann:78}
Herrmann, D.J.: {An old problem for the new psycho-semantics: Synonymity}.
  Psychological Bulletin  85(3),  490--512 (1978)

\bibitem{Heylen:08}
Heylen, K., Peirsman, Y., Geeraerts, D., Speelman, D.: {Modelling Word
  Similarity: an Evaluation of Automatic Synonymy Extraction Algorithms}. In:
  Proceedings of the Sixth International Conference on Language Resources and
  Evaluation. pp. 3243--3249. LREC 2008, European Language Resources
  Association, Marrakech, Morocco (2008)

\bibitem{Hope:13}
Hope, D., Keller, B.: {MaxMax: A Graph-Based Soft Clustering Algorithm Applied
  to Word Sense Induction}. In: Computational Linguistics and Intelligent Text
  Processing: 14th International Conference, CICLing 2013, Samos, Greece, March
  24-30, 2013, Proceedings, Part I, pp. 368--381. Springer Berlin Heidelberg,
  Berlin, Heidelberg (2013)

\bibitem{Lappin:94}
Lappin, S., Leass, H.J.: {An Algorithm for Pronominal Anaphora Resolution}.
  Computational Linguistics  20(4),  535--561 (1994)

\bibitem{Loukachevitch:11}
Loukachevitch, N.V.: {Thesauri in information retrieval tasks}. Moscow
  University Press, Moscow, Russia (2011), in Russian.

\bibitem{Loukachevitch:16}
Loukachevitch, N.V., Lashevich, G., Gerasimova, A.A., Ivanov, V.V., Dobrov,
  B.V.: {Creating Russian WordNet by Conversion}. In: Computational Linguistics
  and Intellectual Technologies: papers from the Annual conference
  ``Dialogue''. pp. 405--415. RSUH, Moscow, Russia (2016)

\bibitem{Manandhar:10}
Manandhar, S., Klapaftis, I., Dligach, D., Pradhan, S.: {SemEval-2010 Task 14:
  Word Sense Induction \& Disambiguation}. In: Proceedings of the 5th
  International Workshop on Semantic Evaluation. pp. 63--68. Association for
  Computational Linguistics, Uppsala, Sweden (2010)

\bibitem{Meyer:12}
Meyer, C.M., Gurevyich, I.: {OntoWiktionary: Constructing an Ontology from the
  Collaborative Online Dictionary Wiktionary}, pp. 131--161. IGI Global,
  Hershey, PA, USA (2012)

\bibitem{Mikolov:13}
Mikolov, T., Sutskever, I., Chen, K., Corrado, G.S., Dean, J.: {Distributed
  Representations of Words and Phrases and their Compositionality}. In:
  Advances in Neural Information Processing Systems 26, pp. 3111--3119. Curran
  Associates, Inc., Harrahs and Harveys, NV, USA (2013)

\bibitem{Navigli:12}
Navigli, R.: {A Quick Tour of Word Sense Disambiguation, Induction and Related
  Approaches}. In: Proceedings of the 38th International Conference on Current
  Trends in Theory and Practice of Computer Science, pp. 115--129. SOFSEM'12,
  Springer-Verlag, Berlin, Heidelberg (2012)

\bibitem{Panchenko:12:cdud}
Panchenko, A., Adeykin, S., Romanov, A., Romanov, P.: {Extraction of Semantic
  Relations between Concepts with KNN Algorithms on Wikipedia}. In: Proceedings
  of the 2nd International Workshop on Concept Discovery in Unstructured Data.
  pp. 78--86. No. 871 in CEUR Workshop Proceedings, Leuven, Belgium (2012)

\bibitem{Panchenko:12:konvens}
Panchenko, A., Morozova, O., Naets, H.: {A Semantic Similarity Measure Based on
  Lexico-Syntactic Patterns}. In: Proceedings of KONVENS 2012. pp. 174--178.
  \"{O}GAI (2012)

\bibitem{Panchenko:16}
Panchenko, A., Simon, J., Riedl, M., Biemann, C.: {Noun Sense Induction and
  Disambiguation using Graph-Based Distributional Semantics}. In: Proceedings
  of the 13th Conference on Natural Language Processing. pp. 192--202. KONVENS
  2016, Bochumer Linguistische Arbeitsberichte (2016)

\bibitem{Panchenko:17}
Panchenko, A., Ustalov, D., Arefyev, N., Paperno, D., Konstantinova, N.,
  Loukachevitch, N., Biemann, C.: {Human and Machine Judgements for Russian
  Semantic Relatedness}, pp. 221--235. Springer International Publishing, Cham,
  Switzerland (2017)

\bibitem{panchenko2011comparison}
Panchenko, A.: Comparison of the baseline knowledge-, corpus-, and web-based
  similarity measures for semantic relations extraction. In: Proceedings of the
  GEMS 2011 Workshop on GEometrical Models of Natural Language Semantics. pp.
  11--21. Association for Computational Linguistics, Edinburgh, UK (2011)

\bibitem{Peirsman:08}
Peirsman, Y., Heylen, K., Speelman, D.: {Putting things in order. First and
  second order context models for the calculation of semantic similarity}. In:
  Proceedings of the 9th Journ{\'e}es internationales d'Analyse statistique des
  Donn{\'e}es Textuelles. pp. 907--916. JADT 2008, Lyon, France (2008)

\bibitem{Pelevina:16}
Pelevina, M., Arefyev, N., Biemann, C., Panchenko, A.: {Making Sense of Word
  Embeddings}. In: Proceedings of the 1st Workshop on Representation Learning
  for NLP. pp. 174--183. Association for Computational Linguistics, Berlin,
  Germany (2016)

\bibitem{Seitner:16}
Seitner, J., Bizer, C., Eckert, K., Faralli, S., Meusel, R., Paulheim, H.,
  Ponzetto, S.P.: {A Large Database of Hypernymy Relations Extracted from the
  Web}. In: Proceedings of the Tenth International Conference on Language
  Resources and Evaluation. pp. 360--367. LREC 2016, European Language
  Resources Association (ELRA), Portoro\v{z}, Slovenia (2016)

\bibitem{Shwartz:16}
Shwartz, V., Goldberg, Y., Dagan, I.: {Improving Hypernymy Detection with an
  Integrated Path-based and Distributional Method}. In: Proceedings of the 54th
  Annual Meeting of the Association for Computational Linguistics (Volume 1:
  Long Papers). pp. 2389--2398. Association for Computational Linguistics,
  Berlin, Germany (2016)

\bibitem{Snow:04}
Snow, R., Jurafsky, D., Ng, A.Y.: {Learning Syntactic Patterns for Automatic
  Hypernym Discovery}. In: Proceedings of the 17th International Conference on
  Neural Information Processing Systems. pp. 1297--1304. NIPS'04, MIT Press,
  Vancouver, British Columbia, Canada (2004)

\bibitem{Ustalov:17:acl}
Ustalov, D., Panchenko, A., Biemann, C.: {Watset: Automatic Induction of
  Synsets from a Graph of Synonyms}. In: Proceedings of the 55th Annual Meeting
  of the Association for Computational Linguistics (Volume 1: Long Papers). pp.
  1579--1590. Association for Computational Linguistics, Vancouver, Canada
  (2017)

\bibitem{Wandmacher:05}
Wandmacher, T.: {How semantic is Latent Semantic Analysis?} In: Proceedings of
  R\'{E}CITAL 2005. pp. 525--534. Dourdan, France (2005)

\bibitem{zeng2007semantic}
Zeng, X.M.: Semantic relationships between contextual synonyms. ERIC  4(9),
  33--37 (2007)

\bibitem{zipf1935psychobiology}
Zipf, G.K.: The psycho-biology off language. Boston: I-Ioughton-Mifflin  (1935)

\end{thebibliography}

\end{document}